\documentclass[10pt]{article}

\PassOptionsToPackage{numbers,sort&compress}{natbib}
\usepackage[dblblindworkshop,final]{neurips_2026}
\workshoptitle{NeurIPS 2026 Workshop on SLMs for Agentic Systems, Paris, France, 2026}

\makeatletter
\renewcommand{\@noticestring}{Accepted in NeurIPS 2026 Workshop on SLMs for Agentic Systems, Paris, France, 2026.}
\makeatother

\usepackage[utf8]{inputenc}
\usepackage[T1]{fontenc}
\usepackage{booktabs}
\usepackage{graphicx}   
\usepackage{float}     
\usepackage{fancyvrb}  
\usepackage{textcomp}

\DeclareUnicodeCharacter{2192}{\textrightarrow}
\DeclareUnicodeCharacter{2014}{---}
\usepackage[expansion=false]{microtype}
\usepackage{xcolor}
\usepackage{amsmath}
\usepackage{url}
\usepackage[colorlinks=true,allcolors=blue]{hyperref}

\newcommand{\rev}[1]{\textcolor{black}{#1}}

\newcommand{\nCases}{280}
\newcommand{\nModels}{9}
\newcommand{\nRecords}{2520}
\newcommand{\accHosted}{95.4}
\newcommand{\accLocalBest}{91.8}
\newcommand{\accGenHosted}{93.9}
\newcommand{\accSingleLocal}{90.7}

\newcommand{\costGenHosted}{2.15}
\newcommand{\savingGenHosted}{82}
\newcommand{\gapLocalBest}{3.6}
\newcommand{\gapGenHosted}{1.4}
\newcommand{\actuatorCostShare}{81}
\newcommand{\actuatorMeanTokens}{28\,122}
\newcommand{\hostedPlannerAcc}{84}
\newcommand{\hostedInfeasible}{20}

\newcommand{\maxOverAct}{87.2}
\newcommand{\maxOverActModel}{Gemma4 E2B}
\newcommand{\alwaysRefuseModel}{Qwen3.5 0.8B}

\newcommand{\eCases}{46}
\newcommand{\eConfigs}{3}

\newcommand{\eHostedDirect}{25/26}

\newcommand{\eHostedAttr}{39/43}

\newcommand{\eGenDirect}{24/26}
\newcommand{\eGenPipe}{15/20}
\newcommand{\eGenAttr}{39/43}

\newcommand{\eLocalDirect}{24/26}
\newcommand{\eLocalPipe}{12/20}

\newcommand{\eHostedSpend}{2.47}
\newcommand{\eGenSpend}{0.70}
\newcommand{\eGenSpendShare}{28}
\newcommand{\eSystemLimited}{3}

\newcommand{\accHaiku}{89.6}

\newcommand{\accGenHostedHaiku}{92.5}
\newcommand{\costGenHostedHaiku}{0.56}

\newcommand{\haikuIntent}{80.6}

\newcommand{\pCodegenSonnet}{0.002}
\newcommand{\pCodegenHaiku}{0.039}
\newcommand{\pActuatorSonnet}{0.062}

\title{Not Every Call Needs a Frontier Model:\\
Per-Call-Site Evaluation of Small Language Models\\
in a Deployed Agentic Home-Automation System}

\author{%
  Panagiotis Kasnesis\textsuperscript{1,2,3} \quad
  Christos Chatzigeorgiou\textsuperscript{1,2} \\
  Lazaros Toumanidis\textsuperscript{1,2} \quad
  Amalia Contiero Syropoulou\textsuperscript{1,3} \\[0.6em]
  \textsuperscript{1}University of West Attica, Athens, Greece \\
  \textsuperscript{2}Waldiez PC \quad
  \textsuperscript{3}ThinGenious PC \\[0.4em]
  \texttt{\{pkasnesis, chrihatz, laztoum, acontiero\}@uniwa.gr}
}

\begin{document}
\maketitle

\begin{abstract}
An agentic system issues several structurally different kinds of LLM calls. It routes intent,
classifies actions, grounds language in a device registry, plans multi-agent pipelines and writes
the Python code those pipelines run. The difficulty of these call sites varies by an order of
magnitude, yet in practice a single model, chosen for the hardest site, serves all of them. In this
work, we evaluate \nModels{} models from 0.8B to a frontier hosted model across the five call sites
of a deployed open-source home-automation framework (Wactorz), using its unmodified production
prompts and two real Home Assistant installations (\nCases{} cases, \nRecords{} scored calls). We
find that capability is not ordered the same way at every site, and that larger models are not
uniformly better: one 4B model is worse than its 2B sibling at grounded actuation. Paired testing
shows the best local model to be statistically indistinguishable from both hosted models at four of
five sites. Only code generation separates them, against a small hosted model
($p=\pCodegenHaiku{}$) as well as a frontier one ($p=\pCodegenSonnet{}$). Aggregate accuracy also
\emph{hides a safety failure specific to actuation}, where small models resolve the
accuracy/refusal trade-off in degenerate ways: one model (\maxOverActModel{}) actuates on
\maxOverAct{}\% of requests for devices the site does not own, while another refuses every request
it receives. Routing each site to its best local model reaches \accLocalBest{}\% against
\accHosted{}\% at no per-call cost. In a live deployment judged by a user, hosting only the two
generative sites matches hosting everything (\eGenAttr{} against \eHostedAttr{}) for
\eGenSpendShare{}\% of the spend, and the actuation gap the benchmark predicted appears as exactly
one case in twenty-six. Benchmark, harness and all records are released at
\url{https://github.com/waldiez/slm-callsite-eval}.
\end{abstract}

\section{Introduction}
\label{sec:intro}

The case for small language models in agentic systems is usually made at the level of a whole
model: can a 4B model serve as the agent? This framing obscures where the practical leverage lies.
A running framework issues several structurally different LLM calls per turn, and their difficulty
varies widely. Deciding whether ``turn off the tv'' is an actuation or an automation request is a
four-way classification with a ten-token budget, while writing an asynchronous agent that
subscribes to an Pub/Sub (e.g., MQTT) topic with a correctly-shaped callback is program synthesis against an
unfamiliar API. Systems typically serve both from one model sized for the second task, and as a
result pay hosted prices for the first, which is also the call made most often.

We treat model selection as a \emph{per-call-site} decision and measure it in a system (i.e., Wactorz~\citep{wactorz}, an open-source multi-agent home-automation runtime in which
long-lived agents are supervised over MQTT) that is
deployed rather than simulated. The first three of its five call sites are evaluated against the framework's production system
prompts, imported from the agent code, so the numbers describe the deployed system rather than a
proxy; all five prompts are reproduced verbatim in Appendix~\ref{app:prompts}.

\paragraph{Contributions.} We contribute (i) a \nCases{}-case benchmark over five call sites of a
real agentic system, grounded in two installations with different inventories, including
cross-site cases answerable at one site and impossible at the other; (ii) a \nModels{}-model
measurement with paired testing, showing per-site orderings that differ, are non-monotonic in
size, and leave only one call site separating local from hosted; (iii) an actuation-safety metric
that separates models usable to drive hardware from models with the same aggregate score that are
not; and (iv) routing policies tested end to end on the live deployment, where the offline
prediction held and three failures appeared that no per-call benchmark can see. Code, benchmark
and all \nRecords{} records are released.\footnote{\url{https://github.com/waldiez/slm-callsite-eval}}

\section{Related work}
\label{sec:related}
\paragraph{Small models in agentic systems.}
\citet{belcak2025slm} argue that most agentic calls are repetitive and narrow, that small models
therefore suffice, and sketch a conversion procedure: instrument a running system, find the calls a
smaller model can serve, and migrate them. This paper carries that procedure out on a system that
is already deployed, and reports what the position paper could only project. Capability becomes a
per-site table whose ordering differs by site and is not monotone in size; reliability becomes the
finding that at actuation the failure is one of calibration rather than accuracy, in both
directions; and the economic illustration becomes a measured \eGenSpendShare{}\% of hosted spend at
no loss in task success. Surveys and on-device work provide the capability picture the position
assumes \citep{lu2024slmsurvey,liu2024mobilellm}.

\paragraph{Routing per query, or per call site.}
Cost-aware serving is otherwise studied \emph{per query}, through cascades that escalate until an
answer is good enough \citep{chen2023frugalgpt} and learned routers that pick a strong or weak
model \citep{ong2024routellm,ding2024hybridllm,sakota2024forc,hu2024routerbench}. All of these
need a judge or router at run time, and all of them can mis-route. Routing in this paper is
instead \emph{per call site}: the framework knows statically which call it is about to make, so
the assignment is a configuration value that needs no extra model and \rev{cannot send a call to
the wrong site}, at the cost of not adapting to a hard instance of an easy site. \rev{The two are
complementary rather than competing: a per-query router could still choose among models
\emph{within} a site.} Agent evaluation asks how good a model is
\emph{at being} an agent \citep{liu2024agentbench,yao2024taubench,patil2025bfcl}, building on the
agent-construction and API-grounding lines
\citep{yao2023react,schick2023toolformer,patil2024gorilla,qin2024toolllm}; we ask instead which
calls of an agent system that already exists can move to a smaller model without the system
getting worse. 

\paragraph{Substitution, not hallucination.}
Closest to our actuator site, \citet{li2025homebench} score valid and invalid
smart-home instructions. Our invalid requests name devices that \emph{the other installation
really owns}, so a wrong answer is substitution rather than hallucination, and it actuates real
hardware. \citet{king2024sasha} map under-specified goals onto device plans. Neither work
decomposes a deployed system by call site, and neither reports cost.

\section{Call sites, benchmark and setup}
\label{sec:setup}

\begin{figure}[t]
\centering
\includegraphics[width=0.7\linewidth]{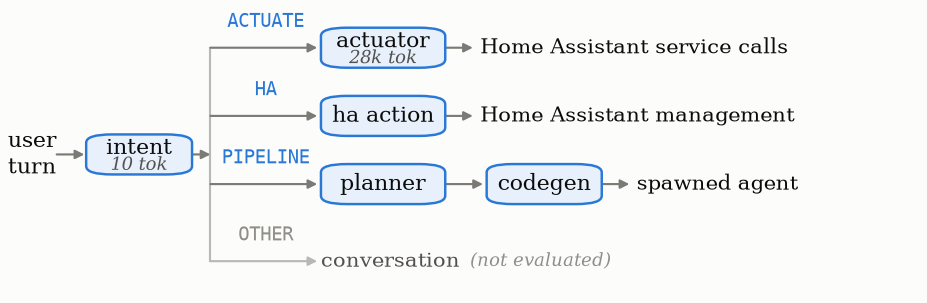}
\caption{The five call sites of one user turn. \emph{Intent} is a static gate:
it runs on every turn under a ten-token budget and fixes which downstream site
executes, so a per-call-site model assignment is a configuration value and
cannot mis-route \rev{between sites}. \emph{Actuator} is a leaf but the expensive one; its prompt
carries the site's whole entity registry, \actuatorMeanTokens{} input tokens on
average, which is where \actuatorCostShare{}\% of hosted spend goes.
\emph{Planner} and \emph{codegen} form the only chain, which is what makes
``hosted for generation only'' a clean cut. The \texttt{OTHER} branch is
ordinary conversation and is not evaluated here.}
\label{fig:callsites}
\end{figure}

\paragraph{Call sites.} Figure~\ref{fig:callsites} shows the five call sites. \rev{The call sites are those of Wactorz's deployed architecture, which we evaluate as it stands rather than propose as a design.} \emph{Intent}
is a four-way label under a ten-token budget. \emph{HA action} classifies a Home
Assistant request into one of the framework's action labels. \emph{Actuator}
maps language to a JSON array of service calls against the site's live entity
registry. \emph{Planner} emits a pipeline specification that must carry concrete
entity ids, topics and schedules rather than describe them, and \emph{codegen}
writes the Python body of each agent that specification names, against the
framework's agent API. Only the last two are generative; the first three emit a
label or a short structured object, which is the asymmetry the rest of this
paper turns on.

\paragraph{Scoring.} Intent and HA action are exact label matches. Actuator requires an
\emph{exact action-set} match: an unrequested extra actuation counts as a failure, because in a
house an unwanted action is worse than none. Cases whose expected output is empty require refusal,
since the device does not exist at that site. Planner requires a valid plan of allowed agent types
that resolves the entity ids and topics it was given rather than describing them. Codegen requires
code that compiles and defines the required \texttt{async} functions; \S\ref{sec:e2e} tightens
this to code that \emph{runs}.

\paragraph{Benchmark.} \nCases{} cases: 36 intent, 28 HA action, 116 actuator, 50 planner, 50
codegen. The actuator set spans two live installations with different inventories\rev{, a household and an
office,} and includes
\emph{flip} cases (a device only the other site owns), \emph{nearmiss} cases (a plausible name
for an absent device) and refusals; sub-groups are encoded in case ids and broken out in
Appendix~\ref{app:subgroups}.

\paragraph{Models and hardware.} \nModels{} models: Qwen3.5 at 0.8B/2B/4B
\citep{qwen2026qwen35}, Gemma4 at E2B/E4B/12B/26B \citep{gemmateam2026gemma4}, and
Claude Sonnet 5 and Haiku 4.5 as hosted references. Local models run on one NVIDIA DGX
Spark (GB10 Blackwell, 128\,GB unified memory), served by llama.cpp
\citep{gerganov2023llamacpp} behind a LiteLLM proxy, with every model at 4-bit GGUF and a
128k context, so no prompt in the suite is truncated. Two properties of these builds
bear on how Table~\ref{tab:main} should be read. First, Gemma4 26B is a mixture of experts
with roughly 4B parameters active per token, so the column order is not an order of
inference cost. Second, the Gemma4 builds are quantization-aware trained while the Qwen3.5
builds are quantized after training, a confound we cannot separate from the architectures.
Exact builds are in Appendix~\ref{app:models}. All calls run at temperature 0, one pass per
case (see \S\ref{sec:limits}).

\section{Results}
\label{sec:results}


\begin{table}[t]
\centering
\footnotesize
\caption{Accuracy (\%) per call site, one pass over the frozen benchmark at temperature 0. Local models are served over an OpenAI-compatible endpoint and cost nothing per call; the hosted rows are measured API spend for the whole 280-case sweep. Paired significance against the best local model is in Table~\ref{tab:mcnemar}.}
\label{tab:main}
\begin{tabular}{lrrrrrrr}
\toprule
Model & Intent & HA action & Actuator & Planner & Codegen & All & Cost \\
 & $n{=}36$ & $n{=}28$ & $n{=}116$ & $n{=}50$ & $n{=}50$ & $n{=}280$ & (\$) \\
\midrule
Qwen3.5 0.8B & 50.0 & 71.4 & 33.6 & 8.0 & 24.0 & \textbf{33.2} & 0 \\
Qwen3.5 2B & 61.1 & 78.6 & 62.1 & 30.0 & 36.0 & \textbf{53.2} & 0 \\
Qwen3.5 4B & 91.7 & 92.9 & 50.9 & 52.0 & 38.0 & \textbf{58.2} & 0 \\
Gemma4 E2B & 83.3 & 89.3 & 56.0 & 46.0 & 34.0 & \textbf{57.1} & 0 \\
Gemma4 E4B & 91.7 & 96.4 & 81.9 & 50.0 & 80.0 & \textbf{78.6} & 0 \\
Gemma4 12B & 86.1 & 89.3 & 95.7 & 76.0 & 84.0 & \textbf{88.2} & 0 \\
Gemma4 26B & 91.7 & 96.4 & 94.8 & 88.0 & 80.0 & \textbf{90.7} & 0 \\
\midrule
Claude Haiku 4.5 & 80.6 & 92.9 & 93.1 & 82.0 & 94.0 & \textbf{89.6} & 3.06 \\
Claude Sonnet 5 & 94.4 & 92.9 & 99.1 & 84.0 & 100.0 & \textbf{95.4} & 12.13 \\
\bottomrule
\end{tabular}
\end{table}

\paragraph{Per-site orderings differ, and are not monotonic in size.}
No model in Table~\ref{tab:main} wins every column: three different local models take the five
sites. Size does not settle the ordering either. Qwen3.5 4B scores below Qwen3.5 2B on actuation
(50.9\% vs.\ 62.1\%), and Gemma4 26B falls below Gemma4 12B on code generation (80.0\% vs.\
84.0\%). Nor is the hosted model uniformly best: Gemma4 26B beats it at planning
(\hostedPlannerAcc{}\% vs.\ 88.0\%), a gap driven by the \emph{infeasible} sub-group, where the
hosted model plans a pipeline for a request the site cannot support \hostedInfeasible{}\% of the
time (Appendix Table~\ref{tab:subplanner}). Scale evidently has not removed this kind of
over-eagerness.

\begin{table}[t]
\centering
\small
\caption{Paired McNemar tests on the same cases, exact two-sided binomial. Each
block compares the best all-round local model (Gemma4 26B) with one hosted
model; \emph{disc.} gives the discordant split as (local only~/~hosted only).
Four of the five call sites are statistically indistinguishable from either
hosted model. Codegen is the exception, and it separates Gemma4 26B from
\emph{both} --- including from the small hosted model, which is why we read the
gap as a property of the call site rather than of frontier scale.}
\label{tab:mcnemar}
\begin{tabular}{lrrrrr}
\toprule
& & \multicolumn{2}{c}{vs.\ Haiku 4.5} & \multicolumn{2}{c}{vs.\ Sonnet 5} \\
\cmidrule(lr){3-4}\cmidrule(lr){5-6}
Call site & Gemma4 26B & disc. & $p$ & disc. & $p$ \\
\midrule
Intent    & 33/36   & 4/0  & 0.125 & 1/2  & 1.000 \\
HA action & 27/28   & 1/0  & 1.000 & 2/1  & 1.000 \\
Actuator  & 110/116 & 6/4  & 0.754 & 0/5  & 0.062 \\
Planner   & 44/50   & 5/2  & 0.453 & 7/5  & 0.774 \\
Codegen   & 40/50   & 1/8  & \textbf{0.039} & 0/10 & \textbf{0.002} \\
\bottomrule
\end{tabular}
\end{table}

\paragraph{Only code generation separates local from hosted.}
Because column-wise gaps invite over-reading, Table~\ref{tab:mcnemar} pairs the best all-round
local model against each hosted model case by case (McNemar exact, \citealp{mcnemar1947}).
On four of five sites it is statistically indistinguishable from both. Code generation separates
it from \emph{both} hosted models ($p=\pCodegenSonnet{}$ against Sonnet 5, $p=\pCodegenHaiku{}$
against Haiku 4.5), with every discordant case favouring the hosted model. The fact that a small,
cheap hosted model reproduces the gap is our strongest evidence that the codegen deficit is a
property of the call site itself rather than of frontier scale. The actuator column remains open:
the comparison misses significance against Sonnet 5 ($p=\pActuatorSonnet{}$), but all five
discordant cases fall in the same direction, which points to insufficient statistical power rather
than a true null result. More sampling passes cannot resolve this, since the test is over a fixed
set of 116 cases.

\paragraph{The codegen gap is concentrated in a handful of API shapes.}
Appendix Table~\ref{tab:subdynamic} splits code generation by the framework API a case
requires, and the deficit turns out to be narrow rather than broad. Gemma4 26B already matches
the hosted models on five of twelve shapes (blocking work, topic contracts, one-shot agents,
publication and state), and loses almost all of its ground on the callback and lifecycle
shapes: \texttt{window} 50, \texttt{poll} 67, \texttt{cleanup} 67 and \texttt{subscribe} 71,
against 100 for both hosted models. These are the same primitives that failed on real
hardware in \S\ref{sec:e2e}, where generated agents called \texttt{agent.window()} and
silently read zero forever. The offline benchmark therefore localised the weakness before the
runtime confirmed it, which supports splitting a call site by the API it exercises instead of
scoring it as a single number. One shape resists both small models: \texttt{llm} defeats
Gemma4 26B and Haiku 4.5 equally (33 each), where Sonnet 5 scores 100.

\paragraph{\rev{What separates codegen from planning.}}
\rev{Both sites are generative, but they ask for different things. The planner emits a
declarative specification over a fixed set of allowed agent types and is scored on whether it
resolves the entity ids and topics it was given, all of which are in its prompt. Codegen must
follow a framework contract that runs against common asynchronous idioms:
\texttt{subscribe} and \texttt{window} are synchronous calls inside async code, callbacks take
exactly one argument, and \texttt{process} is invoked on a poll rather than looping. The prompt
documents these rules with explicit correct and incorrect examples (Appendix~\ref{app:prompts}),
yet the shapes on which local models lose ground are exactly these, while shapes that reduce to
ordinary Python (blocking work, publication, state) are already matched. We therefore read the
gap as difficulty following an unfamiliar, framework-specific API rather than a general limit on
program synthesis at this scale.}

\paragraph{Small hosted is not uniformly better than local.} Haiku 4.5 scores
\haikuIntent{}\% at intent, below five of the seven local models, and \accHaiku{}\% over all five
sites, which is \emph{below} Gemma4 26B's \accSingleLocal{}\%. The two models reach near-identical
totals with opposite profiles: the local model is stronger at routing, and the hosted one is
stronger at code generation.

\paragraph{Aggregate accuracy hides the actuation trade-off.}
Appendix Figure~\ref{fig:safety} splits the actuator site into requests that should produce service
calls and requests that should produce none. Small models fail in two opposite ways. Both are
invisible in a single accuracy number, and both correspond to the calibration failure that
over-refusal benchmarks study in the safety setting \citep{rottger2024xstest,cui2025orbench},
here with a physical consequence. \alwaysRefuseModel{} does not refuse so much as emit a constant: across all 116
actuator cases it returns the same two-token empty array, so its perfect 0.0\%
over-actuation rate and its 0.0\% execution accuracy are the same fact, and the
39 refusals it scores correct carry no evidence that it recognised anything.
It generates normally at every other call site (i.e., 50 distinct outputs at
codegen), so the collapse belongs to this one site; whether that is the
20k-token registry in the prompt or the demand to resolve an exact entity id we
cannot separate, since the two vary together in our design.
\rev{Scale within one family does not settle the direction either: Qwen3.5 2B over-actuates on
66.7\% of impossible requests, while Qwen3.5 4B returns an empty action list on 72.7\% of
actionable ones (Appendix Figure~\ref{fig:safety}). Calibration at this site therefore has to be measured per model rather than inferred from size.}

\paragraph{\rev{Constrained decoding would not close the gap.}}
\rev{A natural mitigation is to restrict the actuator's output to the site's valid entity ids.
That guarantees a well-formed id but not a correct decision to act. Of the 70 service calls that
models emitted on impossible requests, 53 (76\%) already named an entity that exists in the
site's registry: the model chose a real but wrong device, so a grammar over valid ids would not
have blocked them, and for the remainder it would force a choice among real devices unless the
empty action list stays available. Because our impossible requests name devices the
\emph{other} installation owns, the failure is substitution, and preventing it needs calibration
of \emph{whether} to act, not only constraints on \emph{what} may be emitted. Such guardrails remain worth deploying, but alongside a calibrated model rather than in place of one.}

\begin{table}[t]
\centering
\small
\caption{Whole-system accuracy and cost under four routing policies over the same 280 cases. Accuracy is weighted by benchmark case counts, not by a deployment's call mix. Generation-only policies keep intent, HA action and actuation on the best local model per site and send planner and codegen to the hosted model named.}
\label{tab:routing}
\begin{tabular}{lrrrr}
\toprule
Routing configuration & Acc.\ (\%) & \$ total & \$/call & Saving (\%) \\
\midrule
All hosted (Sonnet 5) & 95.4 & 12.13 & 0.0433 & 0 \\
All hosted (Haiku 4.5) & 89.6 & 3.06 & 0.0109 & 75 \\
Hosted for generation only (Sonnet 5) & 93.9 & 2.15 & 0.0077 & 82 \\
Hosted for generation only (Haiku 4.5) & 92.5 & 0.56 & 0.0020 & 95 \\
All local, best per site & 91.8 & 0.00 & 0.0000 & 100 \\
All local, single model (Gemma4 26B) & 90.7 & 0.00 & 0.0000 & 100 \\
\bottomrule
\end{tabular}
\end{table}

\paragraph{Routing beats model choice.}
All-local per-site routing costs nothing and gives up \gapLocalBest{} points against an all-hosted
system; keeping only the generative sites hosted gives up \gapGenHosted{} points and still cuts
spend by \savingGenHosted{}\%\rev{, with the three non-generative sites served by the best local
model for each (i.e., Qwen3.5 4B for intent, Gemma4 E4B for HA action and Gemma4 12B for actuation; a
tie at intent goes to the smallest model)}. Appendix Table~\ref{tab:costprofile} shows why:
\actuatorCostShare{}\% of hosted spend goes to the actuator site, whose prompt carries the entire
entity registry, and this is precisely the site that a mid-sized local model already serves at
95.7\%. In other words, the call that dominates spend is not the call that needs the most capable
model. Which hosted model serves the generative sites is a further lever: routing them to
Haiku 4.5 instead of Sonnet 5 costs \accGenHostedHaiku{}\% against \accGenHosted{}\%, a difference
of 1.4 points, for \$\costGenHostedHaiku{} against \$\costGenHosted{}, a quarter of the spend.
These accuracies are weighted by benchmark case counts rather than by a deployment's call mix; a
traffic-weighted view would favour local routing further.

\section{End-to-end validation on the live deployment}
\label{sec:e2e}

Per-site accuracy indicates which calls a small model can serve, but not whether a system wired
that way works as a whole. We therefore drove the running deployment through its chat endpoint
under three routing configurations on one installation, judging every case as a user would: the
requested effect happened, nothing else was actuated, and the reply claimed nothing untrue.
Judgement is by a person rather than a model \citep{zheng2023judging} or a side-effect check,
because neither of those would catch an agent that registers, runs and does the wrong thing, a
failure that is common here. \eCases{} cases were scored in all \eConfigs{} configurations.

\begin{table}[t]
\centering
\small
\caption{End-to-end outcomes on the live deployment, judged by a person: the requested effect happened, nothing else was actuated, the reply claimed nothing untrue. Only the two generative call sites differ between the middle and right columns. $n{=}46$ cases scored in every configuration, on one installation. 3 cases hit limitations outside the model --- a framework bug in the sliding-window API, and an LLM client a spawned agent could not reach --- and the same 3 are excluded from the second total in every column, so that row does not simply subtract failures: one of the three passed under all-hosted.}
\label{tab:e2e}
\begin{tabular}{lrrr}
\toprule
& all-hosted & gen-hosted & all-local \\
\midrule
Direct (actuate, refuse) & 25/26 & 24/26 & 24/26 \\
Pipeline (scheduled, reactive, spawn) & 15/20 & 15/20 & 12/20 \\
\midrule
\textbf{All cases} & \textbf{40/46} & \textbf{39/46} & \textbf{36/46} \\
Excluding system-limited cases & 39/43 & 39/43 & 35/43 \\
Hosted spend (\$) & 2.47 & 0.70 & 0.00 \\
\bottomrule
\end{tabular}
\end{table}

\paragraph{Hosting only the generative call sites matches hosting everything.}
Sending planner and codegen to the frontier model while keeping everything else local scores
\eGenPipe{} on pipeline cases, the same as hosting every call site, for \eGenSpendShare{}\% of the
spend (\$\eGenSpend{} against \$\eHostedSpend{}). Excluding the \eSystemLimited{} system-limited
cases, the two configurations are indistinguishable at \eGenAttr{}. Running everything locally
costs three pipeline cases (\eLocalPipe{}).

\paragraph{The offline benchmark predicted the deployed outcome.}
Only the generative sites differ between the middle and bottom configurations, so the direct cases
act as a control, and they behave like one: \eGenDirect{} and \eLocalDirect{}, identical across
all 26. Moving actuation from the frontier model to a local one costs exactly one case in
twenty-six (\eHostedDirect{} against \eLocalDirect{}), or four points, and Table~\ref{tab:main}
predicted four (99.1\% against 94.8\%). An offline measurement on a frozen benchmark anticipated
what the deployed system did to real hardware.

\paragraph{Scale is not monotone here either.}
The all-hosted configuration was the only one to fail \emph{in two minutes turn
off the livingroom lights}, classifying it as conversation and actuating nothing, while both
configurations with a local intent classifier handled it. It also answered \emph{when the
livingroom door opens turn on the lights there} by switching on every light in the house, where
the local configurations replied that the site has no door sensor. This is the over-actuation of
\S\ref{sec:results} appearing live, and it came from the model that never over-actuated offline.
Both are cases the cheaper system got right.

\paragraph{What the harness could not have told us.}
Three findings came only from running the system. Two pipeline cases failed because
\texttt{agent.window()} returns a fresh empty buffer on every call, so an agent that counts or
averages reads zero forever; here, correct generated code was defeated by a framework API. A
third case degraded silently to a statistical summary because the spawned agent could not reach an
LLM client. Most consequentially, several requests never reached the codegen site at all: the
framework matches them to a pre-built catalogue recipe and spawns it directly, at no cost and
independent of the routing configuration. This bounds \S\ref{sec:results}: the codegen site
matters less in deployment than a benchmark that weights categories equally would imply.

\section{Limitations}
\label{sec:limits}

One pass per case at temperature 0: the generative sites vary between samples and we report no
variance, though one end-to-end case passed in two configurations and failed in the third, so that
variance is real. Two installations\rev{, a household and an office}, with ground truth annotated by the authors;
\rev{diversity across further homes and buildings (layouts, device mixes, registry sizes) is therefore untested, and
synthetic suites such as HomeBench \citep{li2025homebench} offer a breadth that real
installations cannot, complementing our grounding in live hardware.} The
end-to-end run covers one of them and its verdicts come from a single unblinded rater. \rev{Its
criteria are observable (the effect happened, nothing else moved, the reply claimed nothing
untrue) and the annotations are released for re-judging, while a blinded, multi-rater protocol is future
work.} Three
intent keys and the planner entity block were revised after inspecting outputs. Two of the
affected cases routed to a Home Assistant path the framework has disabled, one asked for a
hardware recommendation that the route's own definition excludes, and the planner prompt omitted
the area labels the deployed system sends, leaving location cases unanswerable. The revisions
moved the hosted model $+0.08$ and three local models $-0.03$, against the direction of our
conclusion, and both key sets are released. Mislabelled answers are common in benchmarks \citep{northcutt2021labelerrors}, and we report
our revisions openly so they can be checked. Sub-group counts are small (2--10), so those
percentages are indicative rather than conclusive. Offline accuracy counts every benchmark case
equally, while a real deployment makes some calls far more often than others; \S\ref{sec:e2e}
shows that this gives the codegen site more weight than it has in practice. Offline
accuracy is weighted by benchmark composition rather than by a real call mix, which
\S\ref{sec:e2e} shows overstates the codegen site. Latency is single-request, with no batching
and no energy measurement\rev{; only wall-clock time per call was recorded, so time to first
token cannot be separated from decoding, and hosted latency includes the network round trip}.
\rev{We did not vary the size of the entity registry, which grows with the home and dominates the
actuator prompt, so how small models degrade as it grows is open. Compound requests that mix an
action with conversation follow the framework's single-label intent contract and are labelled
\texttt{OTHER}, so their actionable part is not evaluated; decomposing them is future work.
Static assignments must be revisited when prompts or models change, and the released harness
turns that into a scripted re-run rather than a manual judgement; we did not compare them against
a learned per-query router. All local models ran on a DGX Spark, a high-end edge device with 128\,GB of unified memory. The 4-bit builds of the mid-sized models have modest memory footprints, but we have not measured them on less powerful edge hardware, where latency and throughput may differ.} Finally, the prompts are one framework's: the call-site
\emph{taxonomy} generalises to agentic systems with a routing/grounding/planning/codegen split,
but the absolute numbers do not.

\section{Conclusion}

Model selection in an agentic system is best treated as a per-call-site routing decision rather
than a single procurement choice. Once the expensive calls are separated from the difficult ones,
most of a deployed framework's traffic can run locally at no per-call cost. For safety, what
matters is not aggregate accuracy but the joint rate of acting when asked and refusing when the
device does not exist; aggregate scores hide this pair, and it decides whether a model can be
trusted with a lock. \rev{The specific assignments belong to one framework and two installations;
what we expect to carry over is the procedure of measuring each call site separately and
checking the result on the live system.} Harness, benchmark, the \nRecords{} offline records and the end-to-end
annotations are available at \url{https://github.com/waldiez/slm-callsite-eval}, and every table and percentage
here is generated from them. All experiments use the production prompts of Wactorz~\citep{wactorz}
v0.5.3, and the live run used a v0.5.3-based deployment.

\begin{ack}
Research supported by the NVIDIA Academic Grant Program using two NVIDIA DGX Spark systems.
\end{ack}

\bibliographystyle{plainnat}
\bibliography{refs}

@article{belcak2025slm,
  title  = {Small Language Models are the Future of Agentic {AI}},
  author = {Belcak, Peter and Heinrich, Greg and Diao, Shizhe and Fu, Yonggan
            and Dong, Xin and Muralidharan, Saurav and Lin, Yingyan Celine
            and Molchanov, Pavlo},
  journal = {arXiv preprint arXiv:2506.02153},
  year   = {2025},
}

@article{chen2023frugalgpt,
  title   = {{FrugalGPT}: How to Use Large Language Models While Reducing Cost
             and Improving Performance},
  author  = {Chen, Lingjiao and Zaharia, Matei and Zou, James},
  journal = {arXiv preprint arXiv:2305.05176},
  year    = {2023},
}

@article{ong2024routellm,
  title   = {{RouteLLM}: Learning to Route {LLM}s with Preference Data},
  author  = {Ong, Isaac and Almahairi, Amjad and Wu, Vincent and Chiang, Wei-Lin
             and Wu, Tianhao and Gonzalez, Joseph E. and Kadous, M. Waleed
             and Stoica, Ion},
  journal = {arXiv preprint arXiv:2406.18665},
  year    = {2024},
}

@inproceedings{patil2025bfcl,
  title     = {The {B}erkeley Function Calling Leaderboard {(BFCL)}: From Tool Use to Agentic Evaluation of Large Language Models},
  author    = {Patil, Shishir G. and Mao, Huanzhi and Yan, Fanjia and Ji, Charlie Cheng-Jie and Suresh, Vishnu and Stoica, Ion and Gonzalez, Joseph E.},
  booktitle = {International Conference on Machine Learning (ICML)},
  year      = {2025}
}

@article{yao2024taubench,
  title   = {$\tau$-bench: A Benchmark for Tool-Agent-User Interaction in
             Real-World Domains},
  author  = {Yao, Shunyu and Shinn, Noah and Razavi, Pedram and Narasimhan, Karthik},
  journal = {arXiv preprint arXiv:2406.12045},
  year    = {2024},
}

@inproceedings{liu2024agentbench,
  title     = {{AgentBench}: Evaluating {LLM}s as Agents},
  author    = {Liu, Xiao and Yu, Hao and Zhang, Hanchen and Xu, Yifan and Lei, Xuanyu and Lai, Hanyu and Gu, Yu and Ding, Hangliang and Men, Kaiwen and Yang, Kejuan and Zhang, Shuntian and Deng, Xiang and Zeng, Aohan and Du, Zhengxiao and Chen, Wei and Shen, Yuxuan and Zhang, Xiaohan and Su, Yu and Dong, Yuxiao and Tang, Jie},
  booktitle = {International Conference on Learning Representations (ICLR)},
  year      = {2024}
}

@article{king2024sasha,
  title     = {Sasha: Creative Goal-Oriented Reasoning in Smart Homes with Large Language Models},
  author    = {King, Evan and Yu, Haoxiang and Lee, Sangsu and Julien, Christine},
  journal   = {Proceedings of the ACM on Interactive, Mobile, Wearable and Ubiquitous Technologies},
  volume    = {8},
  number    = {1},
  pages     = {1--28},
  year      = {2024},
  publisher = {ACM New York, NY, USA},
  doi       = {10.1145/3643505}
}

@inproceedings{li2025homebench,
  title     = {{HomeBench}: Evaluating {LLM}s in Smart Homes with Valid and
               Invalid Instructions Across Single and Multiple Devices},
  author    = {Li, Silin and Guo, Yuhang and Yao, Jiashu and Liu, Zeming and
               Wang, Haifeng},
  booktitle = {Proceedings of the 63rd Annual Meeting of the Association for
               Computational Linguistics (Volume 1: Long Papers)},
  year      = {2025},
  address   = {Vienna, Austria},
}

@misc{qwen2026qwen35,
    title = {Qwen3.5: Towards Native Multimodal Agents},
    url = {https://qwen.ai/blog?id=qwen3.5},
    author = {Qwen Team},
    month = {February},
    year = {2026}
}

@article{gemmateam2026gemma4,
  title   = {Gemma 4 Technical Report},
  author  = {{Gemma Team}},
  journal = {arXiv preprint arXiv:2607.02770},
  year    = {2026},
}

@misc{gerganov2023llamacpp,
  title        = {llama.cpp: {LLM} inference in {C/C++}},
  author       = {Gerganov, Georgi and others},
  year         = {2023},
  howpublished = {\url{https://github.com/ggml-org/llama.cpp}},
}

@article{lu2024slmsurvey,
  title   = {Small Language Models: Survey, Measurements, and Insights},
  author  = {Lu, Zhenyan and Li, Xiang and Cai, Dongqi and Yi, Rongjie and
             Liu, Fangming and Zhang, Xiwen and Lane, Nicholas D. and Xu, Mengwei},
  journal = {arXiv preprint arXiv:2409.15790},
  year    = {2024},
}

@inproceedings{liu2024mobilellm,
  title     = {{MobileLLM}: Optimizing Sub-billion Parameter Language Models
               for On-Device Use Cases},
  author    = {Liu, Zechun and Zhao, Changsheng and Iandola, Forrest and
               Lai, Chen and Tian, Yuandong and Fedorov, Igor and Xiong, Yunyang and
               Chang, Ernie and Shi, Yangyang and Krishnamoorthi, Raghuraman and
               Lai, Liangzhen and Chandra, Vikas},
  booktitle = {International Conference on Machine Learning (ICML)},
  year      = {2024},
}

@inproceedings{ding2024hybridllm,
  title     = {Hybrid {LLM}: Cost-Efficient and Quality-Aware Query Routing},
  author    = {Ding, Dujian and Mallick, Ankur and Wang, Chi and Sim, Robert and
               Mukherjee, Subhabrata and R{\"u}hle, Victor and
               Lakshmanan, Laks V. S. and Awadallah, Ahmed Hassan},
  booktitle = {International Conference on Learning Representations (ICLR)},
  year      = {2024},
}

@article{hu2024routerbench,
  title   = {{RouterBench}: A Benchmark for Multi-{LLM} Routing System},
  author  = {Hu, Qitian Jason and Bieker, Jacob and Li, Xiuyu and Jiang, Nan and
             Keigwin, Benjamin and Ranganath, Gaurav and Keutzer, Kurt and
             Upadhyay, Shriyash Kaustubh},
  journal = {arXiv preprint arXiv:2403.12031},
  year    = {2024},
}

@inproceedings{sakota2024forc,
  title     = {Fly-Swat or Cannon? Cost-Effective Language Model Choice via
               Meta-Modeling},
  author    = {{\v{S}}akota, Marija and Peyrard, Maxime and West, Robert},
  booktitle = {Proceedings of the 17th ACM International Conference on Web
               Search and Data Mining (WSDM)},
  year      = {2024},
}

@inproceedings{rottger2024xstest,
  title     = {{XSTest}: A Test Suite for Identifying Exaggerated Safety
               Behaviours in Large Language Models},
  author    = {R{\"o}ttger, Paul and Kirk, Hannah Rose and Vidgen, Bertie and
               Attanasio, Giuseppe and Bianchi, Federico and Hovy, Dirk},
  booktitle = {Proceedings of the 2024 Conference of the North American Chapter
               of the Association for Computational Linguistics (NAACL)},
  year      = {2024},
}

@inproceedings{cui2025orbench,
  title     = {{OR-Bench}: An Over-Refusal Benchmark for Large Language Models},
  author    = {Cui, Justin and Chiang, Wei-Lin and Stoica, Ion and Hsieh, Cho-Jui},
  booktitle = {Proceedings of the 42nd International Conference on Machine Learning},
  pages     = {11515--11542},
  year      = {2025}
}

@inproceedings{yao2023react,
  title     = {{ReAct}: Synergizing Reasoning and Acting in Language Models},
  author    = {Yao, Shunyu and Zhao, Jeffrey and Yu, Dian and Du, Nan and
               Shafran, Izhak and Narasimhan, Karthik and Cao, Yuan},
  booktitle = {International Conference on Learning Representations (ICLR)},
  year      = {2023},
}

@inproceedings{schick2023toolformer,
  title     = {Toolformer: Language Models Can Teach Themselves to Use Tools},
  author    = {Schick, Timo and Dwivedi-Yu, Jane and Dess{\`i}, Roberto and
               Raileanu, Roberta and Lomeli, Maria and Hambro, Eric and
               Zettlemoyer, Luke and Cancedda, Nicola and Scialom, Thomas},
  booktitle = {Advances in Neural Information Processing Systems (NeurIPS)},
  year      = {2023},
}

@inproceedings{patil2024gorilla,
  title     = {Gorilla: Large Language Model Connected with Massive {APIs}},
  author    = {Patil, Shishir G. and Zhang, Tianjun and Wang, Xin and
               Gonzalez, Joseph E.},
  booktitle = {Advances in Neural Information Processing Systems (NeurIPS)},
  year      = {2024},
}

@inproceedings{qin2024toolllm,
  title     = {{ToolLLM}: Facilitating Large Language Models to Master 16000+
               Real-world {APIs}},
  author    = {Qin, Yujia and Liang, Shihao and Ye, Yining and Zhu, Kunlun and
               Yan, Lan and Lu, Yaxi and Lin, Yankai and Cong, Xin and
               Tang, Xiangru and Qian, Bill and others},
  booktitle = {International Conference on Learning Representations (ICLR)},
  year      = {2024},
}

@inproceedings{zheng2023judging,
  title     = {Judging {LLM}-as-a-Judge with {MT-Bench} and {Chatbot Arena}},
  author    = {Zheng, Lianmin and Chiang, Wei-Lin and Sheng, Ying and
               Zhuang, Siyuan and Wu, Zhanghao and Zhuang, Yonghao and
               Lin, Zi and Li, Zhuohan and Li, Dacheng and Xing, Eric P. and
               Zhang, Hao and Gonzalez, Joseph E. and Stoica, Ion},
  booktitle = {Advances in Neural Information Processing Systems (NeurIPS),
               Datasets and Benchmarks Track},
  year      = {2023},
}

@article{mcnemar1947,
  title   = {Note on the sampling error of the difference between correlated
             proportions or percentages},
  author  = {McNemar, Quinn},
  journal = {Psychometrika},
  volume  = {12},
  number  = {2},
  pages   = {153--157},
  year    = {1947},
}

@inproceedings{northcutt2021labelerrors,
  title     = {Pervasive Label Errors in Test Sets Destabilize Machine
               Learning Benchmarks},
  author    = {Northcutt, Curtis G. and Athalye, Anish and Mueller, Jonas},
  booktitle = {Advances in Neural Information Processing Systems (NeurIPS),
               Datasets and Benchmarks Track},
  year      = {2021},
}

@misc{wactorz,
  author       = {{Wactorz Contributors}},
  title        = {Wactorz: Actor-Model Multi-Agent Framework},
  year         = {2026},
  howpublished = {\url{https://github.com/waldiez/wactorz}},
  note         = {Version 0.5.3}
}

\clearpage
\appendix
\section{Prompts}
\label{app:prompts}

Every prompt below is generated from the constant the running system uses, not transcribed.
The first three are the deployed prompts unchanged. The last two are condensed for the
harness: the deployed planner and codegen prompts are assembled at call time from the live
entity registry and the running agent inventory and reach several hundred lines, so the
variants here keep the output contract that scoring depends on and drop the situational
material. Where a prompt is completed at call time with site-specific content, that content
is named rather than reproduced.


\subsection{Intent routing}
\begingroup\raggedright\noindent
\texttt{\small agents/prompts/main\_actor\_prompts.py}\\
\textbf{INTENT\_CLASSIFIER\_PROMPT}, 28 lines. Used unmodified by the deployed system.
\par\endgroup

\noindent\textbf{Appended at call time:} The user's turn, and a 10-token output budget.

\begin{Verbatim}[fontsize=\footnotesize,frame=leftline,framesep=3mm,xleftmargin=2mm]
You are a routing classifier for a smart home AI assistant.
Respond with exactly one token: ACTUATE, HA, PIPELINE, or OTHER.

ACTUATE = immediate one-shot device control in Home Assistant:
  - Turn on/off a device right now
  - Set temperature, dim lights, lock/unlock door
  - Open/close covers or blinds right now
  - Any direct command whose whole purpose is immediate device control

HA = Home Assistant management, listing, or automation CRUD:
  - List devices, areas, entities, automations
  - Create/edit/delete a HA automation
  - Query what devices or automations exist

PIPELINE = a reactive rule that should run continuously:
  - 'if X happens then do Y' — any conditional/reactive logic
  - 'when X send me a message/notification'
  - 'whenever X turns on/off do Y'
  - Any rule involving a sensor state change triggering an action or notification
  - Any webcam/camera detection triggering anything
  - Anything involving Discord/Telegram notifications triggered by an event

OTHER = general conversation, coding, questions, or mixed requests.anything not HA or pipeli \
  ne related.

Important:
- Choose ACTUATE only when the entire request is immediate device control.
- If the request mixes device control with non-HA tasks, return OTHER.
- If the request is about automations, listing, discovery, or CRUD, return HA.
\end{Verbatim}

\subsection{Home Assistant action classification}
\begingroup\raggedright\noindent
\texttt{\small agents/prompts/home\_assistant\_prompts.py}\\
\textbf{HA\_ACTION\_CLASSIFICATION\_PROMPT}, 35 lines. Used unmodified by the deployed system.
\par\endgroup

\noindent\textbf{Appended at call time:} The user's turn.

\begin{Verbatim}[fontsize=\footnotesize,frame=leftline,framesep=3mm,xleftmargin=2mm]
Classify a Home Assistant user request.
Output exactly one of these strings — nothing else, no punctuation:

recommend_hardware
create_automation
delete_automation
edit_automation
list_automations
list_areas
list_devices
list_entities
other
unknown

Guidelines:
- recommend_hardware  → user only wants hardware/device suggestions, compatibility info, or  \
  to know what the existing hardware can do
- create_automation   → user wants to create/add/build/make a new automation, even if they a \
  lso mention choosing between existing sensors, lights, or devices
- delete_automation   → user wants to delete/remove/disable an existing automation
- edit_automation     → user wants to update/change/rename/modify an existing automation
- list_automations    → user explicitly asks to list/show/enumerate existing automations
- list_areas          → user explicitly asks to list/show/enumerate areas
- list_devices        → user explicitly asks to list/show/enumerate devices as an inventory
- list_entities       → user explicitly asks to list/show/enumerate entities or entity IDs a \
  s an inventory
- other               → Home Assistant related, but not one of the supported operations above
- unknown             → request is unclear or not Home Assistant related

Decision rule:
- If the user asks you to create/build/add/set up an automation, classify as create_automati \
  on.
- Use recommend_hardware only when the user is asking about hardware feasibility or hardware \
   choices and is not asking you to actually create the automation.
- Use other for Home Assistant status/context questions that need current HA data.
- Use other for existence, count, lookup, or state questions about specific HA devices, sens \
  ors, entities, rooms, or device types.
- "Do I have any thermometers?" is other, not list_devices.
- "What is the state of my thermometer?" is other, not list_entities.
- Use other for camera/snapshot/stream questions, e.g. "show me the kitchen camera", "take a \
   snapshot of the front door", "give me the stream for camera.backyard", "do you have any c \
  ameras?", "can you show a camera snapshot?".
- Use unknown for non-Home-Assistant requests.
\end{Verbatim}

\subsection{Grounded actuation}
\begingroup\raggedright\noindent
\texttt{\small agents/one\_off\_actuator\_agent.py}\\
\textbf{\_RESOLVER\_PROMPT}, 71 lines. Used unmodified by the deployed system.
\par\endgroup

\noindent\textbf{Appended at call time:} The user's turn followed by the site's entire entity registry, one entity id and friendly name per line. That list is what makes this the longest prompt in the system and the one that dominates hosted spend; it is site-specific and is not reproduced here.

\begin{Verbatim}[fontsize=\footnotesize,frame=leftline,framesep=3mm,xleftmargin=2mm]
You are a Home Assistant service-call resolver.

Your task:
- Convert the user's natural-language device control request into one or more Home Assistant \
   service calls.
- Use only entities that exist in the provided Home Assistant device discovery payload.
- Return strict JSON only: an array of action objects.

Action schema:
[
  {
    "domain": "light",
    "service": "turn_on",
    "entity_id": "light.living_room_lamp",
    "service_data": {"brightness_pct": 50}
  }
]

Worked examples:

Request: "dim the living room lamp to 50%" with light.living_room_lamp in the payload:
[
  {
    "domain": "light",
    "service": "turn_on",
    "entity_id": "light.living_room_lamp",
    "service_data": {"brightness_pct": 50}
  }
]

Request: "turn off the TV" with media_player.living_room_tv in the payload:
[
  {
    "domain": "media_player",
    "service": "turn_off",
    "entity_id": "media_player.living_room_tv"
  }
]

Request: "turn off the TV" when NO tv/media_player entity exists in the payload
(only lights, switches, etc.):
[]

Rules:
- Return an array, never an object.
- Use the most specific matching entity_id available.
- The chosen entity MUST be the device the user named. NEVER substitute a
  different device: if the user says "TV" and no TV-like entity exists in the
  payload, return [] — do NOT act on a light, switch, or any other device
  instead. Acting on the wrong device is far worse than doing nothing.
- Return ONLY actions for the devices the user asked about in THIS request.
  Do NOT add extra actions for other devices the user did not mention.
  The worked examples above illustrate the output FORMAT only — never copy
  their actions into your answer.
- If the request is ambiguous or no device matches, return [].
- For multiple commands in one request, return multiple actions.
- Only include service_data keys that are needed.
- Common examples:
  - turn on/off light or switch -> light.turn_on / light.turn_off or switch.turn_on / switch \
    .turn_off
  - set a light color -> light.turn_on with service_data containing rgb_color
  - set heating/thermostat temperature -> climate.set_temperature with {"temperature": number}
  - lock/unlock door -> lock.lock / lock.unlock
  - open/close cover/blinds -> cover.open_cover / cover.close_cover
  - brightness percent -> use {"brightness_pct": number}
- Color requests:
  - For "blue", use {"rgb_color": [0, 0, 255]}.
  - For "pink", use {"rgb_color": [255, 105, 180]}.
  - For "red", use {"rgb_color": [255, 0, 0]}; "green" -> [0, 255, 0]; "purple" -> [128, 0,  \
    128]; "white" -> [255, 255, 255].
  - If the user says just "my light" / "the light" and a color is requested, prefer a color- \
    capable light entity over a color-temperature-only light. Color-capable entities usually \
     have state attributes such as supported_color_modes containing hs, rgb, rgbw, rgbww, or \
     xy, or current color_mode hs/rgb/xy.
  - Do not send color service_data to a light that only supports color_temp.
- Do not invent entity IDs.
- Do not return markdown or explanation.
\end{Verbatim}

\subsection{Pipeline planning}
\begingroup\raggedright\noindent
\texttt{\small evalharness.py}\\
\textbf{\_PLANNER\_SYSTEM}, 30 lines. Condensed for the harness; see below.
\par\endgroup

\noindent\textbf{Appended at call time:} The user's turn, the entity registry, and the topics and names of agents already running. The deployed prompt is assembled at call time and runs to several hundred lines; this variant preserves the output contract that scoring depends on.

\begin{Verbatim}[fontsize=\footnotesize,frame=leftline,framesep=3mm,xleftmargin=2mm]
You are designing reactive automation pipelines for a multi-agent IoT system.
Output ONLY a valid JSON array - no explanation, no markdown, no code fences.
Each array item is a COMPLETE spawn config: it must contain the concrete
entity ids, MQTT topics and schedules needed to run, not just a description.

TYPE 1 - "ha_actuator": call a Home Assistant service when an MQTT message arrives
  {"name": "<kebab-case>", "type": "ha_actuator", "description": "<what it does>",
   "mqtt_topics": ["<trigger-topic>"],
   "actions": [{"domain": "<ha-domain>", "service": "<ha-service>",
                "entity_id": "<entity_id from the entity list>", "service_data": {}}],
   "detection_filter": {"<payload-key>": <value>} or null,
   "cooldown_seconds": <number>}

TYPE 2 - "scheduled": fire at a time or interval
  {"name": "<kebab-case>", "type": "scheduled", "description": "<what it fires>",
   "schedule": {"type": "daily", "at": "17:00"} | {"type": "weekly", "at": "07:30",
                "days": ["mon"]} | {"type": "interval", "seconds": 1800},
   "publish_topic": "schedule/<name>/fired"}

TYPE 3 - "dynamic": custom logic in Python
  {"name": "<kebab-case>", "type": "dynamic", "description": "<what it does>",
   "poll_interval": <seconds>, "code": "<async setup/process/handle_task>"}

Rules:
- Use entity ids EXACTLY as they appear in the AVAILABLE HA ENTITIES list.
- Reuse a topic already published by an agent in AVAILABLE AGENTS rather than
  inventing a new one, and use that topic's exact payload field names.
- Delegate to an existing agent by name when one already does the job.
- HA state changes arrive on homeassistant/state_changes/#; filter by entity_id
  inside the payload (state is nested: payload['new_state']['state']).
\end{Verbatim}

\subsection{Agent code generation}
\begingroup\raggedright\noindent
\texttt{\small evalharness.py}\\
\textbf{\_CODEGEN\_SYSTEM}, 34 lines. Condensed for the harness; see below.
\par\endgroup

\noindent\textbf{Appended at call time:} The task description from the plan step being executed. As above, the deployed prompt is longer and assembled at call time.

\begin{Verbatim}[fontsize=\footnotesize,frame=leftline,framesep=3mm,xleftmargin=2mm]
You write Python code for a dynamic agent in a multi-agent IoT framework.
Output ONLY Python code - no explanation, no markdown fences.
Define async functions among: setup(agent), process(agent),
handle_task(agent, payload), cleanup(agent).

AGENT API
  agent.state                        dict, persists across process() calls
  await agent.publish(topic, data)   publish to MQTT
  await agent.mqtt_get(topic)        ONE-SHOT read, returns one payload
  await agent.log(msg)               dashboard log
  await agent.alert(msg, severity)   dashboard alert
  agent.persist(key, value)          save to disk (NOT awaitable)
  agent.recall(key)                  load from disk (NOT awaitable)
  agent.subscribe(topic, callback)   continuous subscription (NOT awaitable)
  agent.window(topic, seconds=N)     sliding window object (NOT awaitable)
  agent.declare_contract(publishes=[...], subscribes=[...])
  await agent.llm.chat(prompt)       the framework's LLM - never import your own

CRITICAL RULES
- agent.subscribe() is NOT awaitable and REQUIRES a callback. It runs as a
  background task; the callback must be an async function taking exactly one
  argument (the payload).
    CORRECT:  async def on_msg(payload): ...
              agent.subscribe('some/topic', on_msg)
    WRONG:    await agent.subscribe('some/topic', on_msg)   # not awaitable
    WRONG:    agent.subscribe('some/topic')                 # no callback
    WRONG:    async def on_msg(topic, payload): ...         # too many args
- agent.window() is synchronous: `w = agent.window(t, seconds=300)`, never await.
- agent.persist() / agent.recall() are synchronous - do not await them.
- process() is called repeatedly on a poll interval. Do NOT write `while True`
  loops inside it; just do one iteration of work and return.
- Import libraries INSIDE functions, never at module level.
- Wrap blocking calls (cv2, torch, psutil-heavy work) in
  `await asyncio.get_event_loop().run_in_executor(None, fn)`.
\end{Verbatim}

\section{Acting versus refusing}
\label{app:safety}
\begin{figure}[H]
\centering
\includegraphics[width=0.62\linewidth]{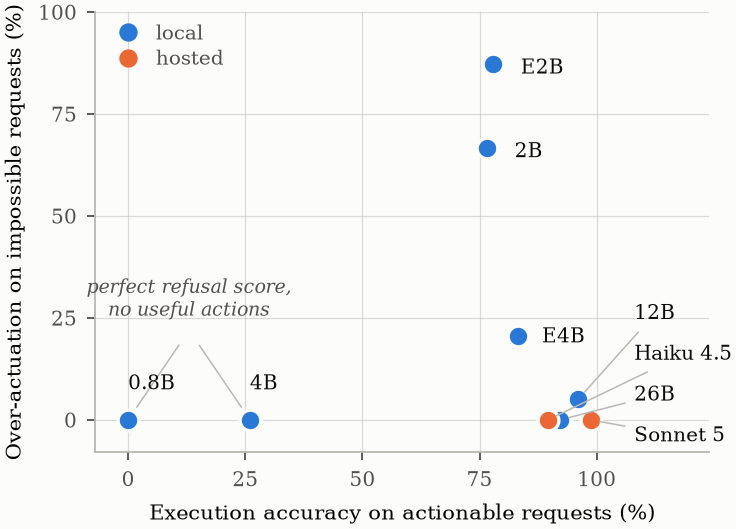}
\caption{Acting versus refusing on the actuator call site ($n{=}77$ actionable,
$n{=}39$ impossible). Each point is one model: execution accuracy on requests
that should produce service calls, against the share of impossible requests that
produced at least one call on real hardware. Deployable is the bottom right;
the bottom left is a model that never actuates and so scores a perfect
over-actuation rate, which is why neither axis is meaningful alone. Qwen3.5 2B
and 4B fail in opposite directions and land at almost the same joint score: 2B
never abstains but actuates on two thirds of the impossible requests, while 4B
returns the empty action list on 56 of the 77 actionable requests and is correct
on 20 of the 21 it answers.}
\label{fig:safety}
\end{figure}

\section{Sub-group breakdowns}
\label{app:subgroups}

\begin{table}[H]
\centering
\small
\caption{Actuator accuracy by cross-site sub-group. Prefixes \emph{a\_}/\emph{b\_} are the two installations; \emph{control} is a device the site owns, \emph{flip} one only the \emph{other} site owns, \emph{nearmiss} a plausible name for an absent device, \emph{same} a device both own, \emph{refusal} a device neither owns.}
\label{tab:subactuatorsite}
\resizebox{\textwidth}{!}{%
\begin{tabular}{lrrrrrrrrrr}
\toprule
Model & a\_control & a\_flip & a\_nearmiss & a\_refusal & a\_same & b\_control & b\_flip & b\_nearmiss & b\_refusal & b\_same \\
$n$ & 28 & 7 & 6 & 6 & 3 & 8 & 7 & 6 & 6 & 3 \\
\midrule
Qwen3.5 0.8B & 0 & 29 & 100 & 100 & 0 & 0 & 71 & 100 & 100 & 0 \\
Qwen3.5 2B & 79 & 43 & 0 & 0 & 100 & 88 & 86 & 67 & 83 & 100 \\
Qwen3.5 4B & 36 & 29 & 100 & 100 & 33 & 12 & 71 & 100 & 100 & 33 \\
Gemma4 E2B & 86 & 57 & 0 & 17 & 100 & 88 & 14 & 0 & 33 & 100 \\
Gemma4 E4B & 93 & 71 & 83 & 100 & 100 & 100 & 71 & 67 & 83 & 100 \\
Gemma4 12B & 100 & 100 & 100 & 100 & 100 & 100 & 86 & 100 & 83 & 100 \\
Gemma4 26B & 100 & 86 & 100 & 100 & 100 & 100 & 100 & 100 & 100 & 100 \\
\midrule
Claude Haiku 4.5 & 96 & 100 & 100 & 100 & 100 & 100 & 100 & 100 & 100 & 100 \\
Claude Sonnet 5 & 100 & 86 & 100 & 100 & 100 & 100 & 100 & 100 & 100 & 100 \\
\bottomrule
\end{tabular}
}
\end{table}

\begin{table}[H]
\centering
\small
\caption{Actuator accuracy by how the device was named: by \emph{name}, by \emph{location}, by both, by an \emph{attribute} (``the lamp on the small table''), by the \emph{switch} it hangs off, or as a \emph{multi}-device request. \emph{refusal} cases name a device the site does not own.}
\label{tab:subactuatornaming}
\begin{tabular}{lrrrrrrr}
\toprule
Model & attribute & location & multi & name & name+location & refusal & switch \\
$n$ & 8 & 5 & 2 & 6 & 4 & 8 & 3 \\
\midrule
Qwen3.5 0.8B & 0 & 0 & 0 & 0 & 0 & 100 & 0 \\
Qwen3.5 2B & 62 & 80 & 100 & 67 & 100 & 0 & 0 \\
Qwen3.5 4B & 38 & 0 & 50 & 17 & 25 & 100 & 33 \\
Gemma4 E2B & 50 & 40 & 50 & 100 & 100 & 25 & 33 \\
Gemma4 E4B & 62 & 60 & 0 & 83 & 75 & 75 & 100 \\
Gemma4 12B & 88 & 80 & 50 & 100 & 100 & 100 & 100 \\
Gemma4 26B & 75 & 100 & 50 & 83 & 100 & 100 & 67 \\
\midrule
Claude Haiku 4.5 & 62 & 40 & 50 & 100 & 100 & 100 & 100 \\
Claude Sonnet 5 & 100 & 100 & 100 & 100 & 100 & 100 & 100 \\
\bottomrule
\end{tabular}
\end{table}

\begin{table}[H]
\centering
\small
\caption{Planner accuracy by sub-group. \emph{infeasible} asks for a pipeline the site cannot support and is scored on refusing to plan one.}
\label{tab:subplanner}
\begin{tabular}{lrrrrrrrr}
\toprule
Model & chain & create & entity & infeasible & reuseagent & reusetopic & schedule & topic \\
$n$ & 4 & 6 & 8 & 5 & 7 & 10 & 5 & 5 \\
\midrule
Qwen3.5 0.8B & 0 & 0 & 12 & 20 & 14 & 10 & 0 & 0 \\
Qwen3.5 2B & 0 & 83 & 38 & 60 & 0 & 30 & 20 & 0 \\
Qwen3.5 4B & 0 & 67 & 62 & 80 & 14 & 80 & 20 & 60 \\
Gemma4 E2B & 25 & 100 & 25 & 80 & 43 & 60 & 20 & 0 \\
Gemma4 E4B & 50 & 100 & 12 & 100 & 43 & 60 & 20 & 20 \\
Gemma4 12B & 25 & 67 & 100 & 40 & 86 & 90 & 100 & 60 \\
Gemma4 26B & 50 & 100 & 100 & 80 & 86 & 100 & 100 & 60 \\
\midrule
Claude Haiku 4.5 & 75 & 100 & 100 & 80 & 71 & 100 & 80 & 20 \\
Claude Sonnet 5 & 100 & 83 & 100 & 20 & 71 & 90 & 100 & 100 \\
\bottomrule
\end{tabular}
\end{table}

\begin{table}[H]
\centering
\small
\caption{Code generation accuracy by required API shape.}
\label{tab:subdynamic}
\resizebox{\textwidth}{!}{%
\begin{tabular}{lrrrrrrrrrrrr}
\toprule
Model & blocking & cleanup & contract & llm & oneshot & poll & publish & schema & state & subscribe & task & window \\
$n$ & 4 & 3 & 3 & 3 & 3 & 3 & 5 & 4 & 5 & 7 & 6 & 4 \\
\midrule
Qwen3.5 0.8B & 25 & 67 & 33 & 0 & 0 & 33 & 40 & 25 & 40 & 0 & 33 & 0 \\
Qwen3.5 2B & 0 & 0 & 33 & 33 & 0 & 67 & 40 & 25 & 80 & 29 & 83 & 0 \\
Qwen3.5 4B & 0 & 0 & 0 & 33 & 0 & 33 & 60 & 75 & 60 & 43 & 83 & 0 \\
Gemma4 E2B & 50 & 33 & 33 & 67 & 67 & 0 & 20 & 0 & 20 & 43 & 67 & 0 \\
Gemma4 E4B & 75 & 67 & 100 & 33 & 100 & 33 & 60 & 100 & 100 & 100 & 100 & 50 \\
Gemma4 12B & 50 & 67 & 100 & 100 & 100 & 33 & 80 & 100 & 100 & 86 & 83 & 100 \\
Gemma4 26B & 100 & 67 & 100 & 33 & 100 & 67 & 100 & 75 & 100 & 71 & 83 & 50 \\
\midrule
Claude Haiku 4.5 & 100 & 100 & 100 & 33 & 100 & 100 & 100 & 100 & 80 & 100 & 100 & 100 \\
Claude Sonnet 5 & 100 & 100 & 100 & 100 & 100 & 100 & 100 & 100 & 100 & 100 & 100 & 100 \\
\bottomrule
\end{tabular}
}
\end{table}

\begin{table}[H]
\centering
\small
\caption{Intent routing accuracy by sub-group.}
\label{tab:subintent}
\begin{tabular}{lrrrrrrrr}
\toprule
Model & crud & delegate & lexical & mixed & multi & query & sched & verbose \\
$n$ & 6 & 6 & 4 & 4 & 2 & 4 & 6 & 4 \\
\midrule
Qwen3.5 0.8B & 17 & 67 & 75 & 50 & 100 & 25 & 33 & 75 \\
Qwen3.5 2B & 33 & 67 & 75 & 25 & 100 & 100 & 33 & 100 \\
Qwen3.5 4B & 83 & 83 & 100 & 100 & 100 & 100 & 83 & 100 \\
Gemma4 E2B & 67 & 67 & 100 & 100 & 100 & 75 & 83 & 100 \\
Gemma4 E4B & 67 & 83 & 100 & 100 & 100 & 100 & 100 & 100 \\
Gemma4 12B & 67 & 67 & 75 & 100 & 100 & 100 & 100 & 100 \\
Gemma4 26B & 83 & 67 & 100 & 100 & 100 & 100 & 100 & 100 \\
\midrule
Claude Haiku 4.5 & 50 & 67 & 75 & 100 & 100 & 75 & 100 & 100 \\
Claude Sonnet 5 & 100 & 83 & 75 & 100 & 100 & 100 & 100 & 100 \\
\bottomrule
\end{tabular}
\end{table}

\begin{table}[H]
\centering
\small
\caption{Home Assistant action classification accuracy by sub-group.}
\label{tab:subha}
\resizebox{\textwidth}{!}{%
\begin{tabular}{lrrrrrrrrr}
\toprule
Model & camera & create & disable & edit & existence & hardware & list & state & unknown \\
$n$ & 3 & 3 & 3 & 3 & 4 & 3 & 4 & 3 & 2 \\
\midrule
Qwen3.5 0.8B & 100 & 67 & 100 & 33 & 50 & 100 & 100 & 67 & 0 \\
Qwen3.5 2B & 100 & 33 & 100 & 67 & 100 & 100 & 75 & 100 & 0 \\
Qwen3.5 4B & 100 & 67 & 67 & 100 & 100 & 100 & 100 & 100 & 100 \\
Gemma4 E2B & 100 & 100 & 100 & 67 & 50 & 100 & 100 & 100 & 100 \\
Gemma4 E4B & 100 & 100 & 100 & 100 & 100 & 100 & 100 & 100 & 50 \\
Gemma4 12B & 100 & 67 & 33 & 100 & 100 & 100 & 100 & 100 & 100 \\
Gemma4 26B & 100 & 67 & 100 & 100 & 100 & 100 & 100 & 100 & 100 \\
\midrule
Claude Haiku 4.5 & 100 & 67 & 67 & 100 & 100 & 100 & 100 & 100 & 100 \\
Claude Sonnet 5 & 100 & 100 & 67 & 100 & 100 & 100 & 100 & 100 & 50 \\
\bottomrule
\end{tabular}
}
\end{table}

\section{Where hosted spend goes}
\label{app:cost}

\begin{table}[H]
\centering
\small
\caption{Where hosted spend goes. The actuator prompt carries the site's full entity list, so one call site dominates the bill for both hosted models --- and it is the site a mid-sized local model already handles. Token counts are provider-reported and are not comparable across model families: the two tokenise identical prompts differently.}
\label{tab:costprofile}
\begin{tabular}{lrrrrrr}
\toprule
& \multicolumn{3}{c}{Claude Sonnet 5} & \multicolumn{3}{c}{Claude Haiku 4.5} \\
\cmidrule(lr){2-4}\cmidrule(lr){5-7}
Call site & Mean in tok. & \$ & Share (\%) & Mean in tok. & \$ & Share (\%) \\
\midrule
Intent & 499 & 0.06 & 0.5 & 348 & 0.01 & 0.5 \\
HA action & 721 & 0.06 & 0.5 & 509 & 0.02 & 0.5 \\
Actuator & 28\,122 & 9.86 & 81.3 & 20\,921 & 2.47 & 80.8 \\
Planner & 9\,703 & 1.73 & 14.3 & 7\,140 & 0.45 & 14.6 \\
Codegen & 814 & 0.42 & 3.5 & 576 & 0.11 & 3.7 \\
\midrule
Total & & 12.13 & & & 3.06 & \\
\bottomrule
\end{tabular}
\end{table}

\section{Model builds and serving}
\label{app:models}

All seven local models run on the same host and the same runtime, so latency differences
in Appendix~\ref{app:latency} reflect the model, not the stack.

\begin{table}[H]
\centering
\footnotesize
\caption{Local model builds. \emph{MTP} is multi-token prediction; \emph{QAT} is
quantization-aware training. Gemma4 26B is A4B: about 4B parameters active per token.}
\label{tab:models}
\begin{tabular}{lllc}
\toprule
Model & Build & Quant. & Notes \\
\midrule
Qwen3.5 0.8B & Qwen3.5-0.8B-UD-Q4\_K\_XL      & Q4 & MTP \\
Qwen3.5 2B   & Qwen3.5-2B-UD-Q4\_K\_XL        & Q4 & MTP \\
Qwen3.5 4B   & Qwen3.5-4B-UD-Q4\_K\_XL        & Q4 & MTP \\
Gemma4 E2B   & Gemma-4-E2B-it-qat-GGUF        & Q4 & MTP, QAT \\
Gemma4 E4B   & Gemma-4-E4B-it-qat-GGUF        & Q4 & MTP, QAT \\
Gemma4 12B   & Gemma-4-12B-QAT-MTP            & Q4 & MTP, QAT \\
Gemma4 26B   & Gemma-4-26B-A4B-it-qat-GGUF    & Q4 & MTP, QAT, MoE \\
\bottomrule
\end{tabular}
\end{table}

Host: NVIDIA DGX Spark, GB10 (Blackwell), 128\,GB unified memory. Runtime: llama.cpp
build 9592, commit \texttt{ac4cddeb0}, fronted by LiteLLM v1.82.6. All builds are the
Unsloth GGUF releases.

\section{Latency and throughput}
\label{app:latency}
\begin{table}[H]
\centering
\small
\caption{Mean wall-clock seconds per call, and aggregate generation throughput. Local latency is measured on the hardware in \S\ref{sec:setup} with no batching; hosted latency includes network round-trip. Throughput aggregates tokens over time rather than averaging per-call rates. All seven local models share one host and one runtime (Appendix~\ref{app:models}), so these differences are the models. Gemma4 26B beats Gemma4 12B because it is a mixture of experts with about 4B parameters active per token, against 12B dense: a larger model that does less work per token.}
\label{tab:latency}
\begin{tabular}{lrrrrrr}
\toprule
Model & Intent & HA action & Actuator & Planner & Codegen & tok/s \\
\midrule
Qwen3.5 0.8B & 0.3 & 0.2 & 2.1 & 2.9 & 3.4 & 122 \\
Qwen3.5 2B & 0.3 & 0.1 & 3.0 & 4.7 & 6.2 & 48 \\
Qwen3.5 4B & 0.5 & 0.3 & 6.0 & 4.3 & 4.9 & 25 \\
Gemma4 E2B & 0.4 & 0.2 & 3.7 & 2.7 & 4.3 & 94 \\
Gemma4 E4B & 0.4 & 0.2 & 7.4 & 3.6 & 2.7 & 28 \\
Gemma4 12B & 0.4 & 0.3 & 14.0 & 9.0 & 3.2 & 14 \\
Gemma4 26B & 0.4 & 0.2 & 10.1 & 5.9 & 2.7 & 21 \\
\midrule
Claude Haiku 4.5 & 0.9 & 0.9 & 2.5 & 3.9 & 2.7 & 64 \\
Claude Sonnet 5 & 2.1 & 2.5 & 3.2 & 5.3 & 4.6 & 43 \\
\bottomrule
\end{tabular}
\end{table}

\end{document}